\documentclass[runningheads]{llncs}
\usepackage[T1]{fontenc}
\usepackage{graphicx}

\usepackage{subfigure}
\usepackage{booktabs}
\usepackage{multirow}
\usepackage{ulem}
\usepackage{amsmath}
\usepackage{amssymb}
\usepackage{bbding}
\begin{document}
\title{FPTN: Fast Pure Transformer Network for Traffic Flow Forecasting}
%
%
\author {Anonymous PAKDD submission}
\author{Junhao Zhang  \and
Junjie Tang \and
Juncheng Jin \and
Zehui Qu\inst{(}\Envelope\inst{)}}
\authorrunning{F. Author et al.}
\institute{Southwest University, Chongqin, China
\email{\{superblack,swu645867768,arieldong522\}@email.swu.edu.cn, quzehui@swu.edu.cn}}
%
\maketitle              
\begin{abstract}
Traffic flow forecasting is challenging due to the intricate spatio-temporal correlations in traffic flow data. Existing Transformer-based methods usually treat traffic flow forecasting as multivariate time series (MTS) forecasting. However, too many sensors can cause a vector with a dimension greater than 800, which is difficult to process without information loss. In addition, these methods design complex mechanisms to capture spatial dependencies in MTS, resulting in slow forecasting speed. To solve the abovementioned problems, we propose a Fast Pure Transformer Network (FPTN) in this paper. First, the traffic flow data are divided into sequences along the sensor dimension instead of the time dimension. Then, to adequately represent complex spatio-temporal correlations, Three types of embeddings are proposed for projecting these vectors into a suitable vector space. After that, to capture the complex spatio-temporal correlations simultaneously in these vectors, we utilize Transformer encoder and stack it with several layers. Extensive experiments are conducted with 4 real-world datasets and 13 baselines, which demonstrate that FPTN outperforms the state-of-the-art on two metrics. Meanwhile, the computational time of FPTN spent is less than a quarter of other state-of-the-art Transformer-based models spent, and the requirements for computing resources are significantly reduced.
\end{abstract}

\section{Introduction}
\label{sec:intro}
Traffic flow forecasting attempts to predict the future traffic flow in road networks based on historical traffic conditions, which have been widely studied in recent years. It plays a significant role in improving the service quality of Intelligent Transportation Systems (ITS) \cite{liSpatialTemporalFusionGraph2021}. Commonly, traffic flow prediction relies on traffic flow sensors distributed in the road network to count the passage of vehicles. The sensor's traffic records have been processed to an equal frequency(e.g., every 5 minutes) sequence over time. Therefore, the data adopted for traffic flow forecasting is spatio-temporal: both have historical traffic flow information and spatial information such as the locations of traffic sensors and structure of the road network \cite{wuGraphWaveNetDeep2019,liuMSDRMultiStepDependency2022}.
 
\begin{figure}[htbp]
    \centering
    \includegraphics[width=0.7\linewidth]{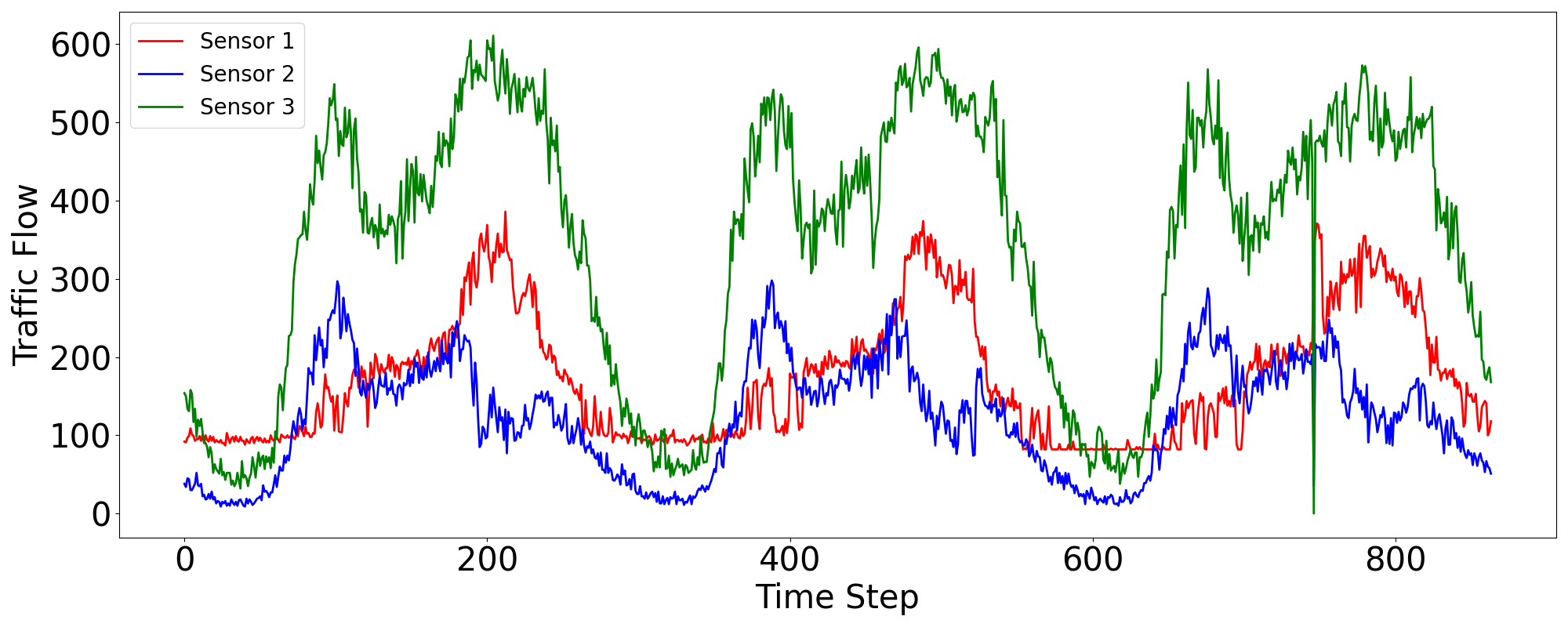}
    \caption{The traffic flow of different sensors has diverse patterns. Sensor 3 has two peaks in one period while Sensor 1 has only one. Sensor 1 and Sensor 2 have different peak times.}
    \label{fig:3senso}
\end{figure}
 
This task is inherently challenging due to the complex and long-range spatio-temporal correlations in traffic networks. As Fig.~\ref{fig:3senso} shows, diversified patterns excited in different traffic sensors, which means the spatio-temporal dependencies are non-linear, dynamic, and shifting long-range \cite{liuMSDRMultiStepDependency2022}. There is a large variety of previous studies for solving this problem. Early traditional works \cite{lippiShortTermTrafficFlow2013} used time series models to capture the correlation by statistical and traditional machine learning methods. However, these approaches only take temporal information into consideration, spatial features have been ignored. Not long after that, deep learning techniques have been widely used to capture prominent spatio-temporal patterns. These approaches typically model temporal dependencies with variants of Recurrent Neural Networks (RNN) \cite{baiPassengerDemandForecasting2019}, and model spatial dependencies with Convolutional Neural Networks (CNN) \cite{zhangDeepSpatioTemporalResidual2017} or Graph Convolutional Networks (GCN) \cite{liDiffusionConvolutionalRecurrent2018}. Some methods \cite{wuGraphWaveNetDeep2019,liSpatialTemporalFusionGraph2021,liuMSDRMultiStepDependency2022} further combined these spatial and temporal models to capture spatio-temporal correlations from traffic data jointly. 
  However, these methods extract temporal features and spatial features separately and then fuse them, which could lose the interlaced spatial and temporal information. Additionally, GCN mainly uses man-made predefined graphs to represent the spatial topological relationship between sensors, which cannot fully portray the complex spatio-temporal dependencies between sensors. 
Recently, Transformer \cite{attenallned} has achieved great success in NLP and CV. Thus researchers began to use Transformer for traffic flow forecasting. However, they regarded traffic flow forecasting as multivariate time series forecasting and designed other complex mechanisms (e.g., GCN) to capture the spatial correlation in the sequence. That results in (1) hard sequence embedding due to a large number of sensors, and the dimension of the vector in the multivariate time series could be too long. (2) Limited spatio-temporal feature extraction in traffic flow data due to GCN. (3) Significant time overhead in making predictions.

To remedy the above problems, we propose a novel framework, termed as Fast Pure Transformer Network (FPTN), which utilizes pure Transformer architecture to extract complex spatio-temporal correlations simultaneously. Specifically, We divide the traffic flow data along the sensor dimension instead of dividing the data along the time dimension as in the previous Transformer-based methods. In this way, a vector in the input sequence represents the historical traffic flow of a sensor. Consequently, we can simultaneously mine the complex spatio-temporal information in the traffic flow data by using the self-attention mechanism in the Transformer encoder without any other complicated mechanism (e.g., GCN). Our model mainly includes two parts: the input embedding and the Transformer encoder. The embedding layer can embed the temporal information of the data, learn the location information of the input sequence, and project the traffic flow data into a suitable vector space for representation. The Transformer encoder uses the self-attention mechanism to extract complex spatio-temporal relationships from sequences partitioned along the sensor dimension. 

The main contributions of this work are concluded as follows:

(1) We divide the traffic flow data along the sensor dimension and utilize the Transformer encoder to extract complex spatio-temporal correlations simultaneously.

(2) We propose the FPTN, which uses pure Transformer architecture to capture spatio-temporal dependencies without using other complex mechanisms, which significantly speeds up the forecasting speed.

(3) Extensive experiments demonstrate that our approach is superior to other state-of-the-art methods.
\section{Related Work}
\subsubsection{Traffic Flow Forecasting.} 
In previous traditional methods, researchers often use statistical time-series models or machine learning models to capture temporal correlations of time series from traffic flow data \cite{zhengGMANGraphMultiAttention2020}. After that, deep learning techniques have been widely adopted. CNN was utilized to capture spatial correlations in STResNet \cite{zhangDeepSpatioTemporalResidual2017}. ConvLSTM \cite{shiConvolutionalLSTMNetwork2015} integrated CNN and LSTM to jointly model spatial and temporal dependencies. Some variants of RNN models combine traffic graphs to capture long-range spatio-temporal dependencies \cite{choiGraphNeuralControlled2022,liuMSDRMultiStepDependency2022}. Recent studies formulate traffic flow prediction on graphs and utilize GCN models for spatio-temporal forecasting \cite{fuTrafficFlowDriven2021,roySSTGNNSimplifiedSpatioTemporal2021,shenTwoTowerSpatialTemporalGraph2022}. 
 GraphWaveNet \cite{wuGraphWaveNetDeep2019a} combined GCN and gated temporal convolution to process traffic conditions. STFGNN \cite{liSpatialTemporalFusionGraph2021} designed a spatio-temporal fusion graph. FOGS \cite{raoFOGSFirstOrderGradient2022} proposed a learning graph to extract spatial correlation adaptively. 
 
\subsubsection{Transformer.} 
Motivated by its strong capability of capturing global spatio-temporal information, Transformer-based models have been applied to traffic flow forecasting tasks \cite{niuMu2ReSTMultiresolutionRecursive2022}. GMAN \cite{zhengGMANGraphMultiAttention2020} proposed spatial and temporal attention mechanisms with gated fusion to capture the complex spatio-temporal dependencies. STTN \cite{xuSpatialTemporalTransformerNetworks2021a} designed spatial Transformer and temporal Transformer to model various scales of spatial dependencies and capture long-range temporal dependencies. Bi-STAT \cite{chen2022bidirectional} proposed a Bidirectional Spatial-Temporal Transformer, which further utilizes the past recollection of past traffic conditions. 

\section{Problem Formulation}
We can represent the traffic network as an undirected graph $\mathcal{R}=(V,E_{road})$, where $\left | V \right | =N$ is the set of sensors, $N$ denotes the number of sensors, and $E$ denotes the road segments between sensors. Denote the observed traffic flow
\begin{math}
    X_\mathcal{R}\in \mathbb{\mathbb{R}}^{N\times T\times C}
\end{math} 
means it presents the observation of traffic network $R$ during $T$ consecutive time steps, whose element is observed $C$ traffic features (e.g., the speed, volume). In this paper, we focus on forecasting traffic volume so that $C=1$, the observed traffic flow can be reshaped as $X_\mathcal{R}\in \mathbb{\mathbb{R}}^{N\times TC}$, and speed and density can be similarly calculated as traffic volume.

\subsubsection{Previous Works.} Previous works treat traffic flow forecasting as multivariable time series forecasting. They denote the observed traffic conditions on $\mathcal{R}$ at the $t$-th time step as a vector 
$X_\mathcal{R}^{(t)} \in \mathbb{R}^N$, 
where the $i$-th element of 
$X_\mathcal{R}^{(t)}$ 
is the traffic condition observed by the $i$-th sensor at the $t$-th time step. The aim of traffic flow forecasting is to find a function $f$ to forecast the next $K$ steps multivariable time series based on the past $T$ steps multivariable time series.
\begin{equation}
    \left ( X_\mathcal{R}^{(t-T+1)},..., X_\mathcal{R}^{(t)} \right ) \stackrel{f}\longrightarrow \left ( X_\mathcal{R}^{(t+1)},..., X_\mathcal{R}^{(t+K)} \right )
\end{equation}

However, it's hard to embed multivariable time series into the Transformer model without information loss when $N$ is large, for example, in PeMSD7 dataset  $N=883$, which means $X_\mathcal{R}^{(t)} \in \mathbb{R}^{883}$, the existing Transformer-based model commonly needs to compress the vector to a suitable dimension such as 256 and then input it into the Transformer.

\subsubsection{Our Work.} We denote the observed traffic conditions on $\mathcal{R}$ at the $n$-th sensor as a vector $X_\mathcal{R}^{(n)} \in \mathbb{R}^{TC}$($C=1$), where the $i$-th element of $X_\mathcal{R}^{(n)}$ is the traffic condition observed by the $n$-th sensor at the $i$-th time step, and $T$ is the number of steps in the observed time interval. Traffic flow forecasting aims to find a function $f$ to forecast the next $K$ steps traffic conditions of $N$ sensors $\hat{X}_\mathcal{R}\in \mathbb{\mathbb{R}}^{N\times KC}$ based on the past $T$ steps traffic conditions of $N$ sensors $X_\mathcal{R}\in \mathbb{\mathbb{R}}^{N\times TC}$
\begin{equation}
    \left ( X_\mathcal{R}^{(1)},X_\mathcal{R}^{(2)},X_\mathcal{R}^{(3)},..., X_\mathcal{R}^{(n)} \right ) \stackrel{f}\longrightarrow \left ( \hat{X}_\mathcal{R}^{(1)},\hat{X}_\mathcal{R}^{(2)},\hat{X}_\mathcal{R}^{(3)},..., \hat{X}_\mathcal{R}^{(n)} \right )
\end{equation}

\section{Methodology}
\begin{figure*}
    \centering
    \includegraphics[width=\textwidth]{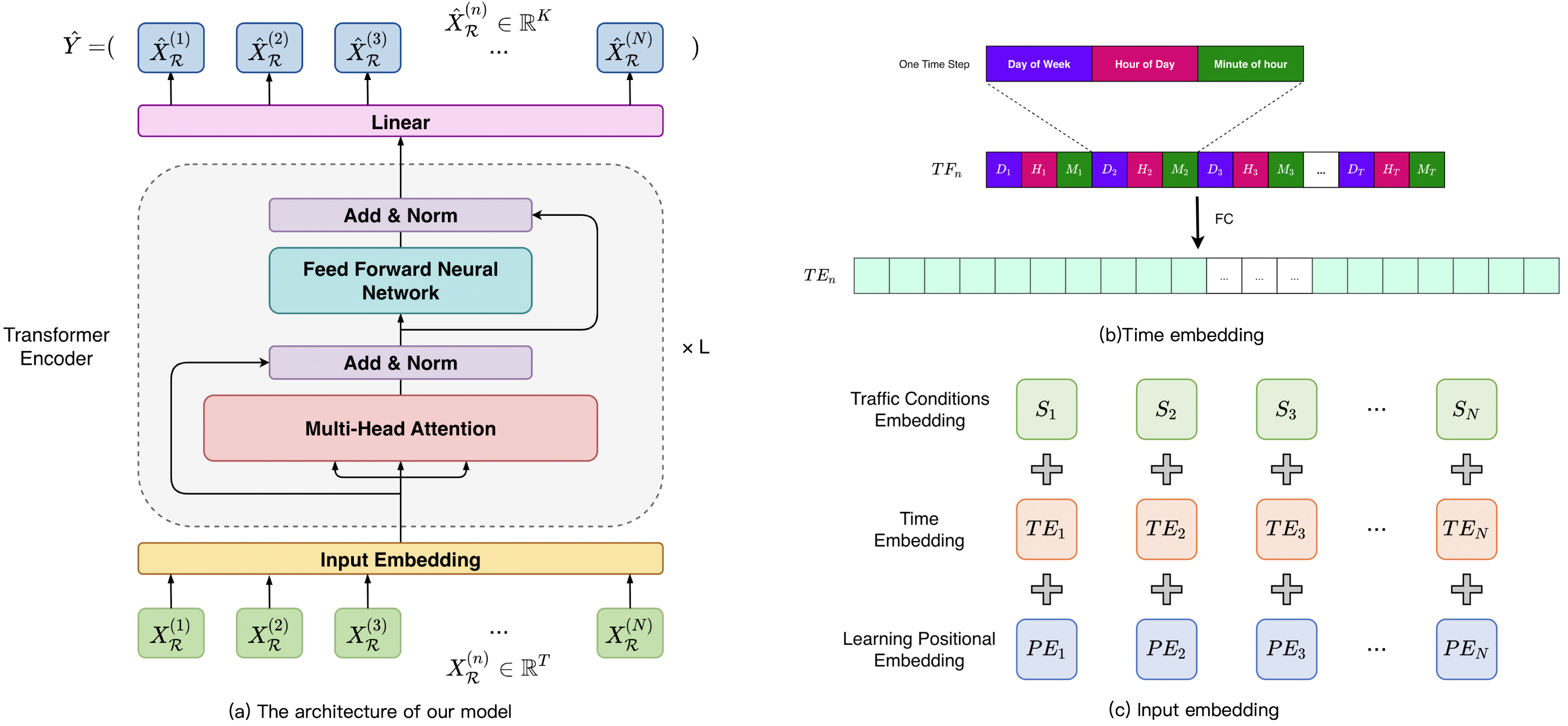}
    \caption{(a) The architecture of the proposed Fast Pure Transformer Network (FPTN). $X_\mathcal{R}^{(n)} \in \mathbb{R}^T$ is the vector of the traffic flow data decomposed along the sensor dimension, which represents the observation value of the traffic conditions in $T$ time steps at the $n$-th sensor. Then $X_\mathcal{R}^{(n)}$ is processed by the (c) input embedding layer and fed into the Transformer Encoder. (b) is the process of time embedding in FPTN.}
    \label{fig:framework}
\end{figure*}
We present the framework of FPTN in Fig.~\ref{fig:framework}. It consists of (1) an input embedding layer, (2) stacked Transformer encoders and (3) an output layer.
\subsection{Input Embedding}
\subsubsection{Traffic Conditions Embedding.} The historical traffic conditions observed by all sensors is $X_\mathcal{R}\in \mathbb{\mathbb{R}}^{N\times TC}$. Thus, in FPTN the traffic conditions is a vector $X_\mathcal{R}^{(n)} \in \mathbb{R}^T$ ($C=1$) for one sensor. To represent the complex spatio-temporal relationship, we project the vector $X_\mathcal{R}^{(n)} \in \mathbb{R}^T$ to the dimension $d_{model}$ and $d_{model} > T$ (e.g., $T=12$ and $d_{model}=256$):
\begin{equation}
    S=X_\mathcal{R}\cdot W^s + b^s
\end{equation}
where $S\in \mathbb{R}^{N\times d_{model}}$ is the embedding of traffic conditions observed by $N$ sensors, $W^s\in \mathbb{R}^{T \times d_{model}}$ and $b^s\in \mathbb{R}^{d_{model}}$ are the weight and bias of traffic conditions embedding.

\subsubsection{Time Embedding.} To represent dynamic correlations among temporal dimensions, we design a time embedding to encode the time step of each sensor's traffic conditions into a vector, as shown in Fig.~\ref{fig:framework}(b). To be specific, we first normalized three time features: $D$ is day-of-week, $H$ is hour-of-day, and $M$ is minute-of-hour into $\mathbb{R}^3$ using min-max scaling for each time step, then concatenate them into a vector $TF_n \in \mathbb{R}^{3T}$ along the temporal dimension. Next, we apply a one-layer fully-connected neural network to transform the time feature $TF_n\in \mathbb{R}^{3T}$ to a time embedding vector $TE_n \in \mathbb{R}^{d_{model}}$:
\begin{equation}
    TE_n=TF_n\cdot W^t+b^t
\end{equation}
$W^t\in \mathbb{R}^{3T\times d_{model}}$ and $b^t\in \mathbb{R}^{d_{model}}$ are the weight and bias of Time Embedding.
Therefore, we obtain the Time Embedding, represented as:
\begin{equation}
    TE=TF\cdot W^t+b^t
\end{equation}
where $TE\in \mathbb{R}^{N\times d_{model}}$, $TF\in \mathbb{R}^{N\times 3T}$.

\subsubsection{Learning Positional Embedding.} 
Unlike NLP and time series tasks, the $N$ vectors in the input sequence of FPTN are not in strict order because they correspond to $N$ sensors. Therefore, we adopted a learning positional embedding instead of using sinusoidal positional encoding, which is fixed. Specifically, we utilize a learnable parameter matrix $PE\in \mathbb{R}^{N\times d_{model}}$ so that the positional embedding for each sensor is $PE_n\in \mathbb{R}^{d_{model}}$. In this way, FPTN can adaptively perform positional embedding during training, let the model learn the position relationship between N vectors by itself, and assist the model in mining the spatial correlation in traffic conditions.

Finally, as shown in Fig.~\ref{fig:framework}(c), we add time embedding $TE\in \mathbb{R}^{N\times d_{model}}$ and learning positional embedding $PE\in \mathbb{R}^{N\times d_{model}}$ to the traffic conditions embedding $S\in \mathbb{R}^{N\times d_{model}}$:
\begin{equation}
    E=S+TE+PE
\end{equation}
$E\in \mathbb{R}^{N\times d_{model}}$ is the input sequence of Transformer encoder.

\subsection{Transformer Encoder}
Since the traffic data is spatio-temporal data \cite{wuGraphWaveNetDeep2019,liuMSDRMultiStepDependency2022}, it is crucial to extract the intertwined and complex spatio-temporal correlations from $E\in \mathbb{R}^{N\times d_{model}}$. In FPTN, we use the Transformer's encoder to capture the spatio-temporal relationships in the traffic data simultaneously. Specifically, the attention mechanism in Transformer's encoder can extract the correlations among each vector in each encoder layer, which means the temporal and spatial correlations among $N$ sensors can be calculated simultaneously. Therefore, our model does not need to design some additional complex mechanisms to extract spatial and temporal correlations separately and then integrate them. Consequently, FPTN can significantly reduce the computational costs of the model with a pure Transformer structure. In addition, FPTN avoids the loss of information caused by separate feature extraction and fusion.

As shown in Fig.~\ref{fig:framework}, the Transformer encoder we used for FPTN is similar to the conventional Transformer \cite{attenallned} model, mainly composed of three parts: multi-head attention, feed-forward neural network, and residual-link \& normalization. Multi-head attention mechanism can learn complex and diverse spatio-temporal patterns in traffic flow data. Formally, the multi-head attention function combines queries $Q$, keys $K$, and values $V$ with $h$ heads as follows:
\begin{equation}
    MultiHead(Q,K,V) = Concat(Head_1,...,Head_h)W^O
\end{equation}
\begin{equation}
    Head_i = Attention(QW_i^Q, KW_i^K, VW_i^V )
\end{equation}
Here $W_i^Q$, $W_i^K$, $W_i^V$ $\in \mathbb{R}^{d_{model}\times \frac{d_{model}}{h}}$, and $W^O\in \mathbb{R}^{d_{model}\times d_{model}}$. In addition, the attention function is scaled dot-product attention as follows:
\begin{equation}
    Attention=softmax(\frac{QK^T}{\sqrt{D} } )V
\end{equation}
Where $D$ is the dimension of $Q$, $K$.

Feed-forward network consists of two linear transformations with a GELU activation in between as follows:
\begin{equation}
FFN\left ( x \right ) =GELU(xW_1+b_1)W_2+b_2
\end{equation}
The dimensionality of input and output is $d_{model}$, and the inner-layer has dimensionality $d_{ff} = 4d_{model}$.

Since the input sample length of the data we use here is fixed and there is no extreme length variation, we use batch normalization instead of layer normalization \cite{zerveasTransformerbasedFrameworkMultivariate2021}.

\subsection{Output Layer}
In FPTN, the input and output vector space is the same, which is the traffic flow data of $N$ sensors. Therefore, our model framework does not use decoder and cross-attention. As shown in Fig.~\ref{fig:framework}(a), we use a fully connected layer network directly after the Transformer encoder to get the output $\hat{X}_\mathcal{R}^{(n)} \in \mathbb{\mathbb{R}}^{K}$:
\begin{equation}
\hat{X}_\mathcal{R}^{(n)} = Z_n\cdot W_o + b_o
\end{equation}
where $Z_n \in \mathbb{R}^{d_{model}}$ is the final vector representations of Transformer encoder, $W_o\in \mathbb{R}^{d_{model\times K}}$, $b_o\in \mathbb{R}^{K}$ are output layer parameters. Finally, the predictions of next $K$ time steps traffic flow denoted as $\hat{Y} = (\hat{X}_\mathcal{R}^{(1)},\hat{X}_\mathcal{R}^{(2)},...,\hat{X}_\mathcal{R}^{(n)}) \in \mathbb{\mathbb{R}}^{N\times K}$.

We choose the mean absolute error (MAE) loss function to train FPTN:
\begin{equation}
\mathcal{L}\left ( \Theta  \right ) =  \frac{1}{N\times K}\sum_{n=1}^{N}\sum_{i=1}^{K} \left | Y_{n,i} - \hat{Y}_{n,i} \right |   
\end{equation}
where $\Theta$ represents all learnable parameters in FPTN, $Y_{n,i}$ is the ground truth of sensor $n$ at time step $i$.

\section{Experiments}
In this section, we describe our experimental environments and results. Our software and hardware environments are as follows: PYTHON 3.9.7, PYTORCH 1.12.0, NUMPY 1.19.2, PANDAS 1.4.3 and CUDA 11.3, Intel(R) Core(TM) i5-11400 CPU @2.6GHz, 32GB RAM and one NVIDIA RTX 3090 GPU, which is a personal computer configuration. We use 4 datasets and 13 baseline models.

\subsection{Database}
In the experiment, we use four real-world traffic datasets, namely PeMSD3, PeMSD4, PeMSD7, and PeMSD8, which come from \cite{liSpatialTemporalFusionGraph2021} and the detailed statistics of four datasets are shown in Table~\ref{tab:dadasets}. 
\begin{table}[]
    \centering
    \caption{The summary of the datasets used in our work} 
    \label{tab:dadasets}
    \begin{tabular}{ccccc}
\hline
Dataset & Sensors & Time Steps & Time Range & Sample Rate \\ \hline
PeMSD3 & 358 & 26,208 & 09/2018 - 11/2018 & 5mins \\
PeMSD4 & 307 & 16,992 & 01/2018 - 02/2018 & 5mins \\
PeMSD7 & 883 & 28,224 & 05/2017 - 08/2017 & 5mins \\
PeMSD8 & 170 & 17,856 & 07/2016 - 08/2016 & 5mins \\ \hline
\end{tabular}
\end{table}

\subsection{Experimental Settings.} All four datasets are aggregated into 5-min windows to generate 12 data points (time steps) per hour and consequently 288 data points per day. We forecast the next $K=12$ steps traffic conditions based on the past $T=12$ steps. Z-score normalization is applied to the traffic data for more stable training. The datasets are split with a ratio of 6:2:2 and 7:1:2 into training, validating, and testing sets. We use the mean absolute error (MAE), the mean absolute percentage error (MAPE), and the root mean squared error (RMSE) to measure the performance of different models, and finally record the best result of our model.

\subsubsection{Baseline.}To evaluate the overall performance of our work, we compare our model with widely used baselines and state-of-the-art models, including: \textbf{ARIM-A}, autoregressive integrated moving average model. 
\textbf{GraphWaveNet} \cite{wuGraphWaveNetDeep2019}. \textbf{STF-GNN} \cite{liSpatialTemporalFusionGraph2021}, spatial-temporal fusion graph neural networks. \textbf{STG-NCDE} \cite{choiGraphNeuralControlled2022}, spatio-temporal graph neural controlled differential equation. \textbf{GMSDR} \cite{liuMSDRMultiStepDependency2022}, graph-based multi-step dependency relation networks. \textbf{FOGS} \cite{raoFOGSFirstOrderGradient2022}, first-order gradient supervision model. \textbf{GMAN} \cite{zhengGMANGraphMultiAttention2020}, graph multi-attention network. \textbf{Traffic Transformer} \cite{caiTrafficTransformerCapturing2020}. \textbf{STTN} \cite{xuSpatialTemporalTransformerNetworks2021a}, spatial-temporal transformer network. \textbf{Bi-STAT} \cite{chen2022bidirectional}, bidirectional spatial-temporal adaptive transformer.

\subsubsection{Hyperparameters.} For FPTN, we validate with the following hyperparameter conﬁgurations: we train for 400 epochs using the RAdam optimizer, with a batch size of 64 on all datasets. The dimension of input embedding $d_{model}$ is in \{$64$, $128$, $256$, $512$, $1024$\}. The heads $h$ of multi-heads attention mechanism is in \{$4$, $8$, $16$, $32$\}. The number of stacked Transformer encoder layers $L$ is in \{$2$, $3$, $4$, $5$, $6$\}. The learning rate $lr$ in all methods is in \{$5\times10^{-3}$, $1\times10^{-3}$, $5\times10^{-4}$, $1\times10^{-4}$\}. An early stop strategy with the patience of 40 iterations on the 
training dataset is used.

\subsection{Experimental Result}

\subsubsection{Computation Time.} Firstly, we compare the computation time of FPTN with AGCRN, DCRNN, GMAN and Bi-STAT on the PeMSD4 dataset in Table~\ref{tab:computime}. It can be observed that AGCRN and DCRNN run slower than the Transformer-based model: GMAN, Bi-STAT and FPTN due to the time-consuming recurrent structure of the RNNs. Moreover, FPTN costs much less computation time than GMAN and Bi-STAT for training and inference owing to the non-decoder Transformer structure of FPTN and no additional processing mechanism (e.g., GCN or RNN). Specifically, the training time of FPTN is 23.62 s/epoch which is 10.31\% of Bi-STAT’s 229.01 s/epoch, the inference time of FPTN is 2.69s which is much faster than Bi-STAT’s 11.56s with ratio 76.73\%.  
\begin{table}
    \centering
    \caption{The computation time on the PeMSD4 dataset.}
    \label{tab:computime}
    \begin{tabular}{ccccc}
\hline
Model & Training(s/epoch) & Inference(s) & MAE & RMSE \\ \hline
DCRNN & 377.09 & 26.78 & 21.22 & 33.44 \\
AGCRN & 249.54 & 25.28 & 19.83 & 32.26 \\
GMAN & 237.31 & 11.01 & 19.36 & 31.06 \\
Bi-STAT & 229.01 & 11.56 & 18.74 & 30.31 \\ \hline
\textbf{FPTN} & \textbf{23.62} & \textbf{2.69} & \textbf{18.49} & \textbf{30.29} \\ \hline
\end{tabular}
\end{table}
\subsubsection{Forecasting Performance Comparison.} The experiment results on the four public datasets are shown in Table~\ref{tab:Result table}. Overall, our proposed method, FPTN, outperforms all baseline methods on all four datasets and two metrics: MAE and RMSE. Specifically, FPTN outperforms FC-LSTM, STSGCN and STFGNN with ratios 33.49\%, 17.81\% and 15.00\% under the MAE metric on PeMSD7 dataset. 
Compared with Transformer-based models GMAN, STTN and Traffic-Transformer, FPTN outperforms GMAN, STTN and Traffic-Transformer with ratios 6.30\%, 10.98\% and 10.98\%, respectively, under the RMSE metric on PeMSD3 dataset. Meanwhile, our model costs far less time than GMAN in training and inference, which have been discussed above. In summary, FPTN achieves better performance and runs much faster than baseline models. The reason might be that we use the pure Transformer architecture to adequately capture the complex spatio-temporal correlations without adding other mechanisms.

FPTN cannot achieve the best performance under the MAPE metric, but it is still comparable with state-of-the-art methods like FOGS and Bi-STAT. We found that is because FPTN is not sensitive when the target value of traffic flow is small (e.g., target value\textless5) which can result in a high value by calculating MAPE.
\begin{table*}[!t]
    \caption{Forecasting error on PeMSD3, PeMSD4, PeMSD7 and PeMSD8. 
    }
    \label{tab:Result table}
    \resizebox{\textwidth}{!}{
    \begin{tabular}{cccc|ccc|ccc|ccc}
    \hline
    \multirow{2}{*}{Model} &  & PeMSD3 &  &  & PeMSD4 &  &  & PeMSD7 &  &  & PeMSD8 &  \\ \cline{2-13} 
 & MAE & RMSE & MAPE & MAE & RMSE & MAPE & MAE & RMSE & MAPE & MAE & RMSE & MAPE \\ \hline
ARIMA & 35.41 & 47.59 & 33.78\% & 33.73 & 48.80 & 24.18\% & 38.17 & 59.27 & 19.46\% & 31.09 & 44.32 & 22.73\% \\
FC-LSTM & 21.33 & 35.11 & 23.33\% & 26.77 & 40.65 & 18.23\% & 29.98 & 45.94 & 13.20\% & 23.09 & 35.17 & 14.99\% \\
GraphWaveNet & 19.12 & 32.77 & 18.89\% & 24.89 & 39.66 & 17.29\% & 26.39 & 41.50 & 11.97\% & 18.28 & 30.05 & 12.15\% \\
STSGCN & 17.48 & 29.21 & 16.78\% & 21.19 & 33.65 & 13.90\% & 24.26 & 39.03 & 10.21\% & 17.13 & 26.80 & 10.96\% \\
AGCRN & 15.98 & 28.25 & 15.23\% & 19.83 & 32.26 & 12.97\% & 22.37 & 36.55 & 9.12\% & 15.95 & 25.22 & 10.09\% \\
STFGNN & 16.77 & 28.34 & 16.30\% & 20.48 & 32.51 & 16.77\% & 23.46 & 36.60 & 9.21\% & 16.94 & 26.25 & 10.60\% \\
GMAN & 16.49 & 26.48 & 17.13\% & 19.36 & 31.06 & 13.55\% & 21.48 & 34.55 & 9.01\% & 14.51 & 23.68 & 9.45\% \\
Traffic-Transformer & 16.39 & 27.87 & 15.84\% & 19.16 & 30.57 & 13.70\% & 23.90 & 36.85 & 10.90\% & 15.37 & 24.21 & 10.09\% \\
STTN & 16.11 & 27.87 & 16.19\% & 19.32 & 30.79 & 13.15\% & 21.05 & {\uline{33.77}} & 8.94\% & 15.28 & 24.25 & 9.98\% \\
STG-NCDE & 15.57 & 27.09 & 15.06\% & 19.21 & 31.09 & 12.76\% & 20.53 & 33.84 & 8.80\% & 15.45 & 24.81 & 9.92\% \\
GMSDR & 15.78 & 26.82 & 15.33\% & — & — & — & — & — & — & 16.36 & 25.58 & 10.28\% \\
FOGS & {\uline {15.13}} & {\uline {24.98}} & {\textbf{14.37\%}} & 19.35 & 31.33 & \uline{12.71\%} & {\uline {20.62}} & 33.96 & \textbf{8.58\%} & 14.92 & 24.09 & \uline{9.42\%} \\
Bi-STAT & 15.29 & 27.54 & 15.19\% & {\uline {18.74}} & {\uline{30.31}} & \textbf{12.59\%} & 20.64 & 34.03 & 8.88\% & {\uline {14.07}} & {\uline {23.45}} & \textbf{9.27\%} \\ \hline
\textbf{FPTN} & \textbf{14.62} & \textbf{24.81} & \uline{14.61\%} & \textbf{18.49} & \textbf{30.29} & 13.10\% & \textbf{19.94} & \textbf{32.49} & \uline{8.77\%} & \textbf{13.98} & \textbf{23.30} & 10.06\% \\ \hline
\end{tabular}
}
\end{table*}

\begin{figure}[htbp]
    \centering
    \subfigure[Node 146 in PeMSD3]{
        \label{fig:PeMSD4_curve}
        \begin{minipage}[t]{0.3\linewidth}
        \centering
        \includegraphics[width=\textwidth]{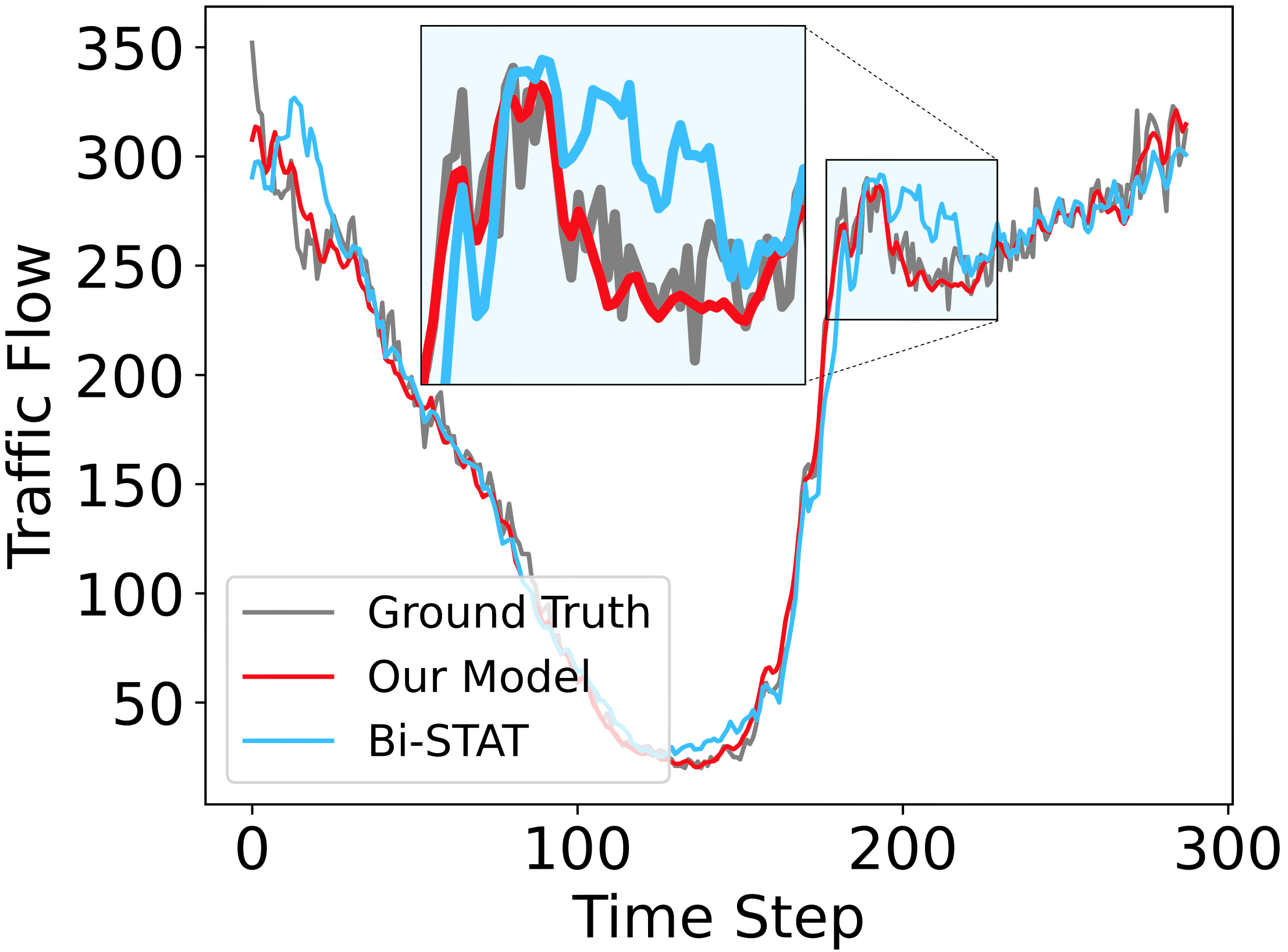}
        \end{minipage}
    }
    \subfigure[Node 869 in PeMSD7]{
        \label{fig:PeMSD7_curve}
        \begin{minipage}[t]{0.3\linewidth}
        \centering
        \includegraphics[width=\textwidth]{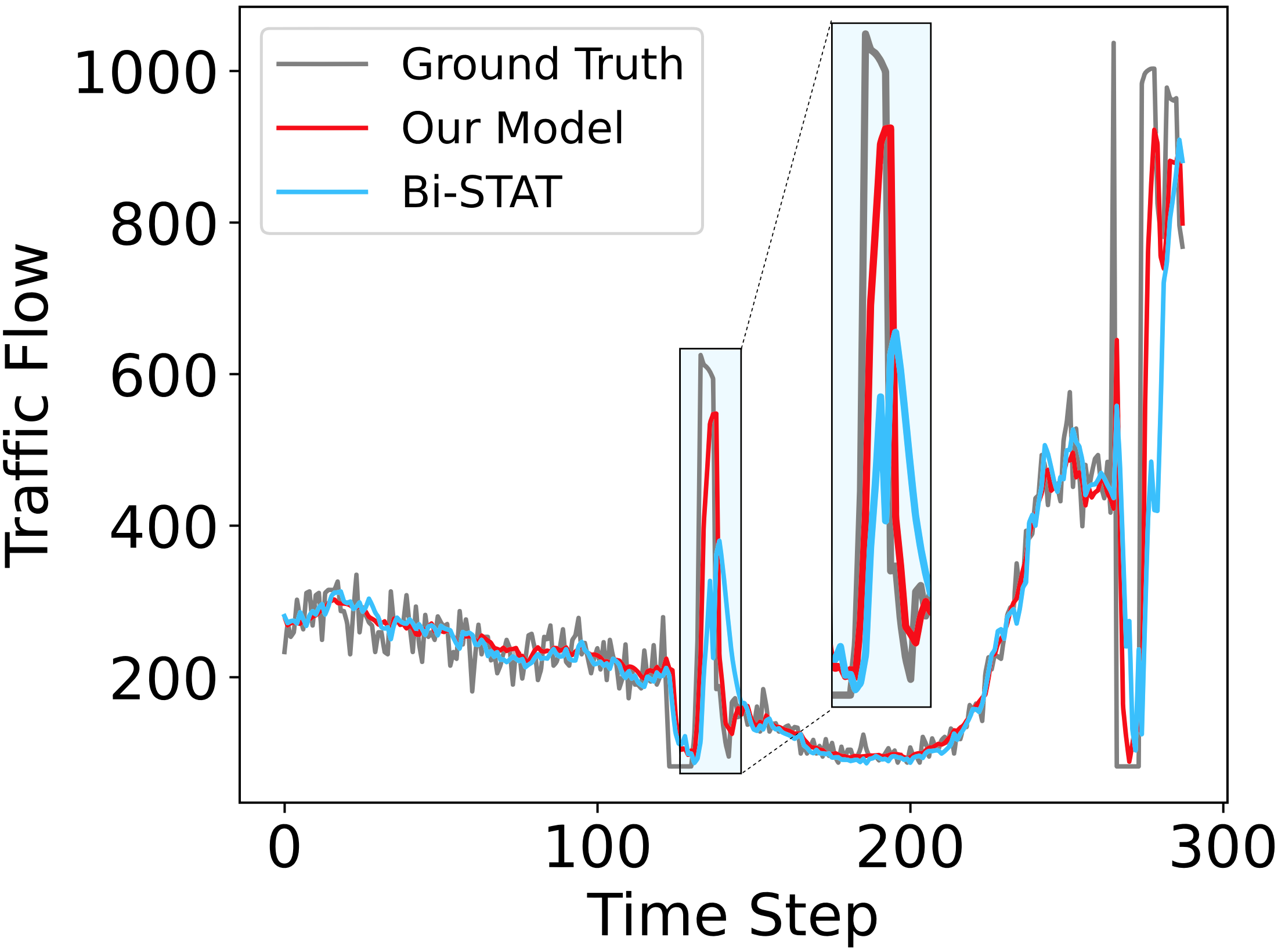}
        \end{minipage}
    }
    \subfigure[Node 78 in PeMSD8]{
        \label{fig:PeMSD8_curve}
        \begin{minipage}[t]{0.3\linewidth}
        \centering
        \includegraphics[width=\textwidth]{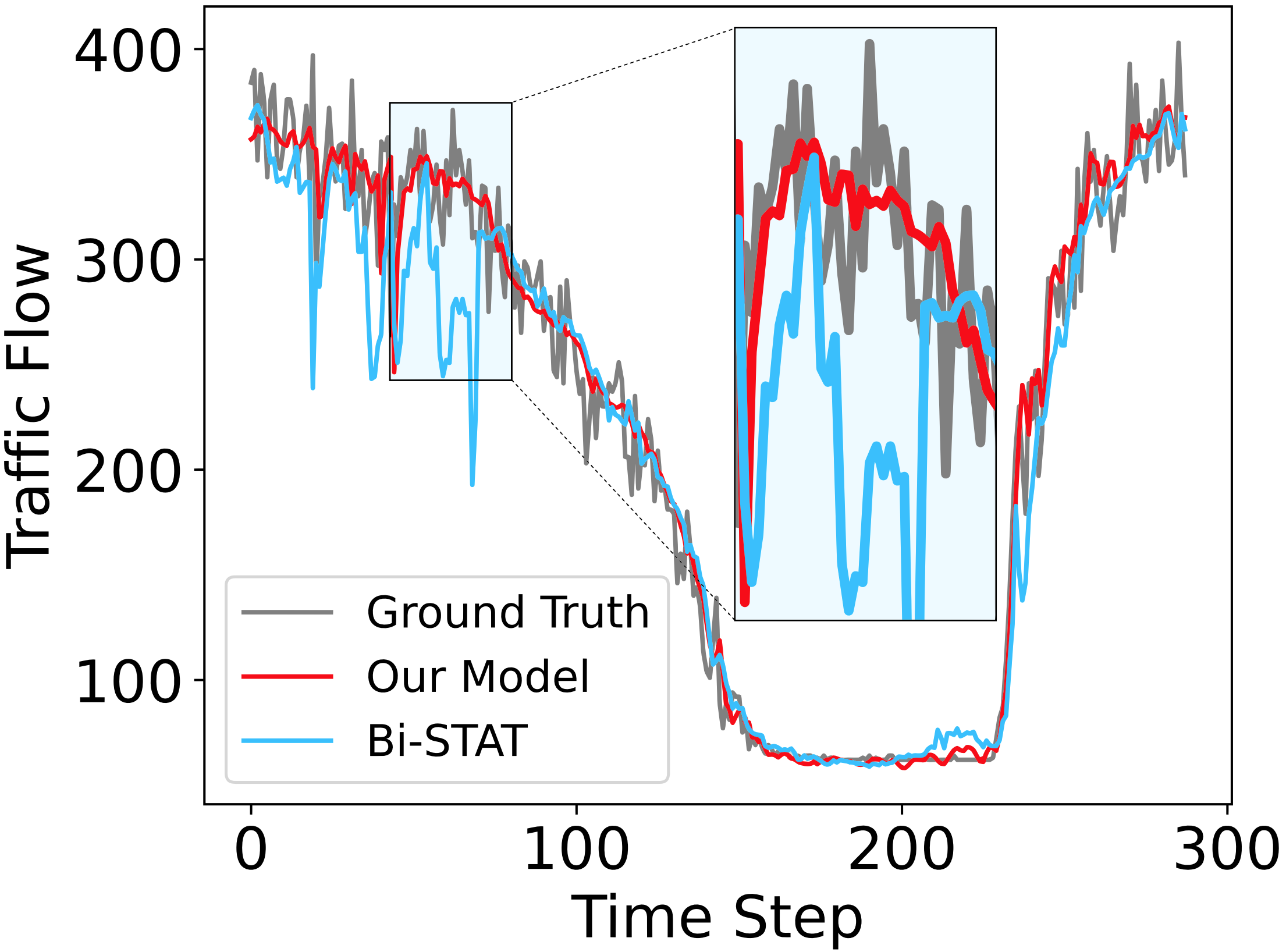}
        \end{minipage}
    }
    
    \caption{Predicted trafﬁc ﬂow visualization on PeMSD3, PeMSD7 and PeMSD8 }
    \label{fig: fit_curve}
\end{figure}

We also visualize the ground-truth and the predicted curves by our model and Bi-STAT in Fig~\ref{fig: fit_curve}. We can see that our model can accurately generate the prediction sequences even at the high traffic areas (Node 869 in PeMSD7) in Fig~\ref{fig: fit_curve}(c). Since Bi-STAT shows reasonable performance, its predicted curve is similar to that of our model in many time steps. As highlighted with boxes, however, our model shows much more accurate predictions for challenging cases(e.g., highlighted time-points for Node 146 in PeMSD3 and Node 78 in PeMSD8).

\subsubsection{Parameter Analysis.} 
\begin{figure*}[htbp]
    \centering
    \subfigure[$d_{model}$]{
        \label{fig:d_model}
        \begin{minipage}[t]{0.3\linewidth}
        \centering
        \includegraphics[width=\textwidth]{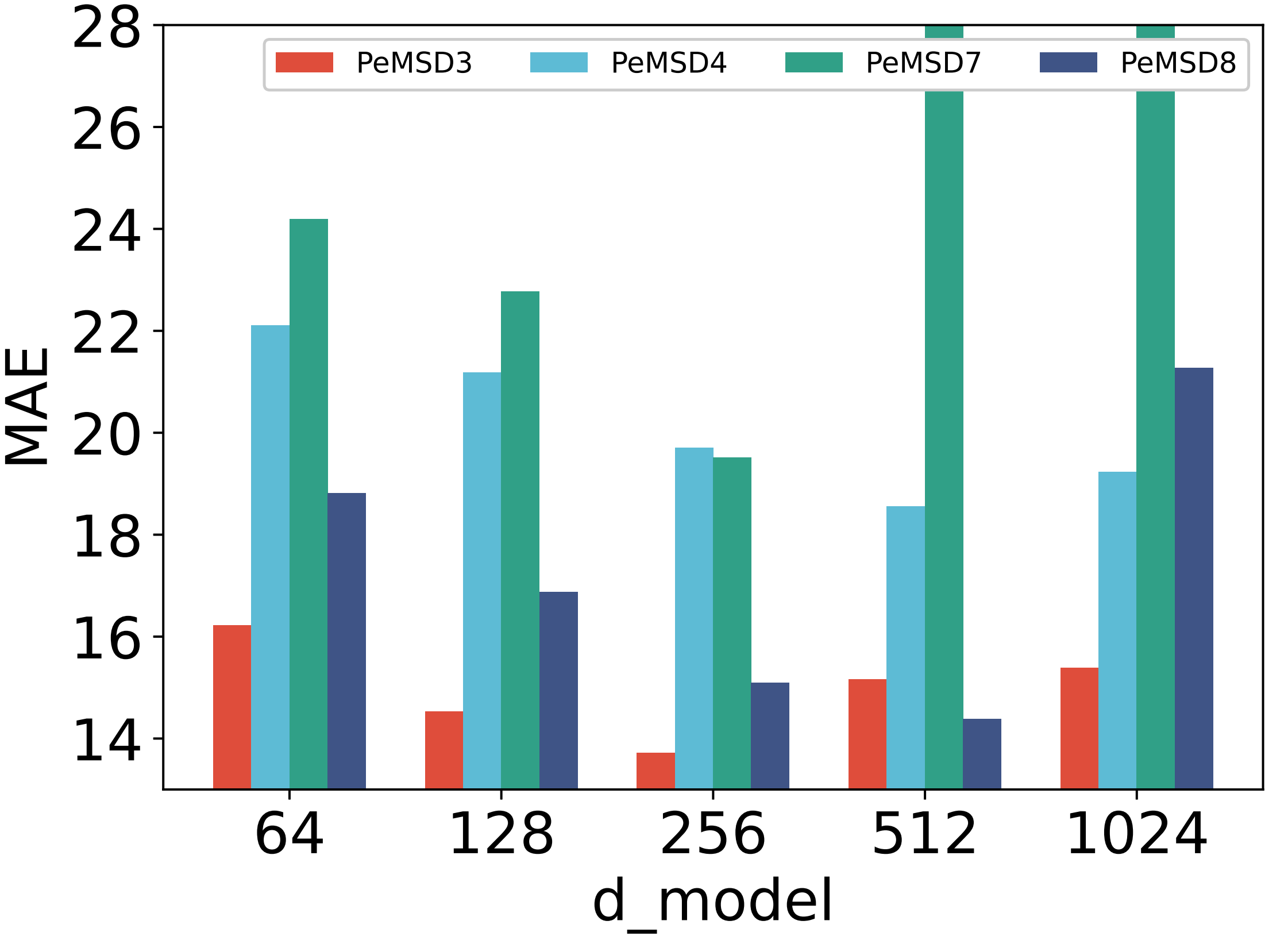}
        \end{minipage}
    }
    \subfigure[$L$]{
        \label{fig:Layer}
        \begin{minipage}[t]{0.3\linewidth}  
        \centering
        \includegraphics[width=\textwidth]{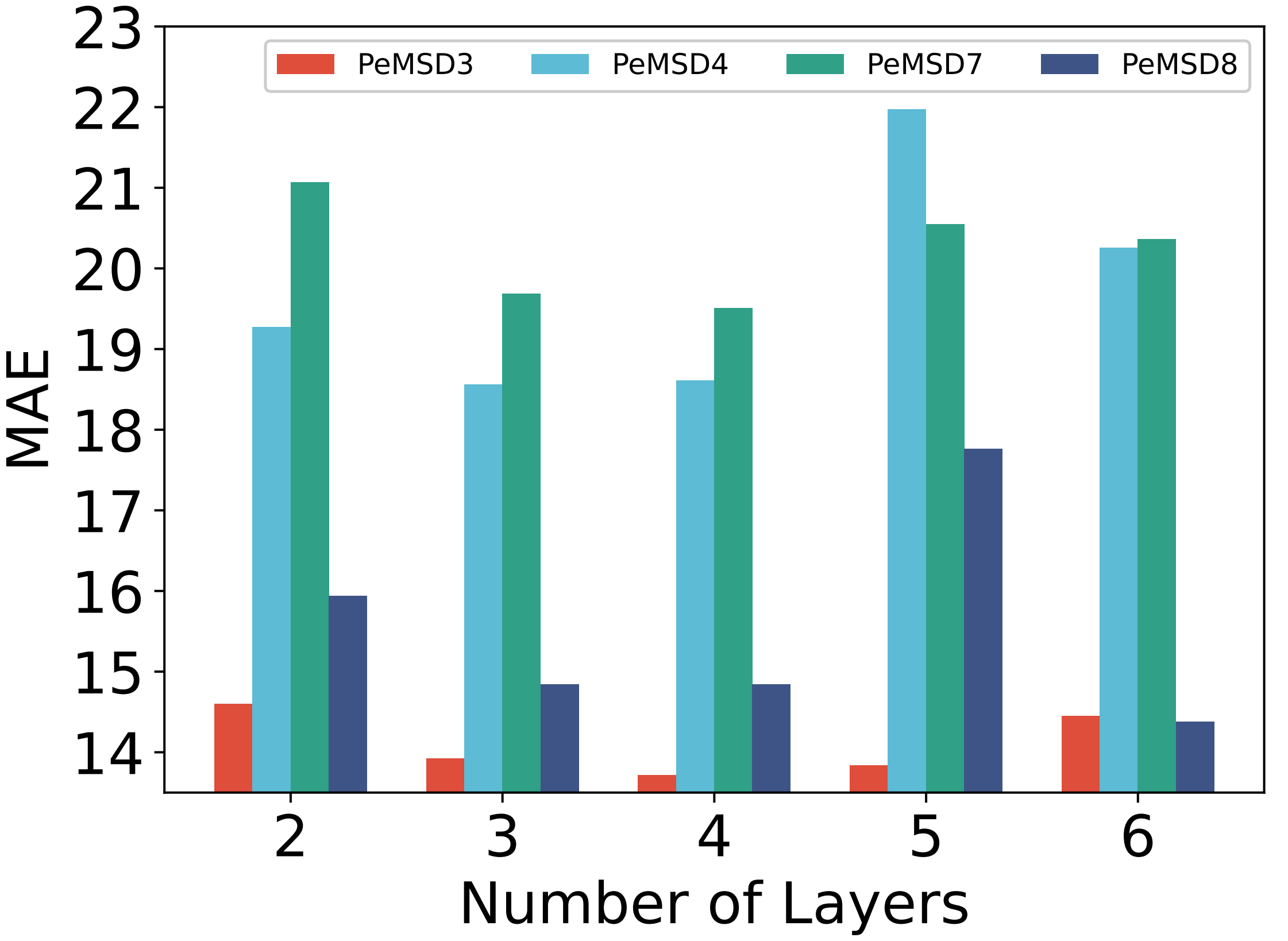}
        \end{minipage}
    }
    \subfigure[$h$]{
        \label{fig:head}
        \begin{minipage}[t]{0.3\linewidth}
        \centering
        \includegraphics[width=\textwidth]{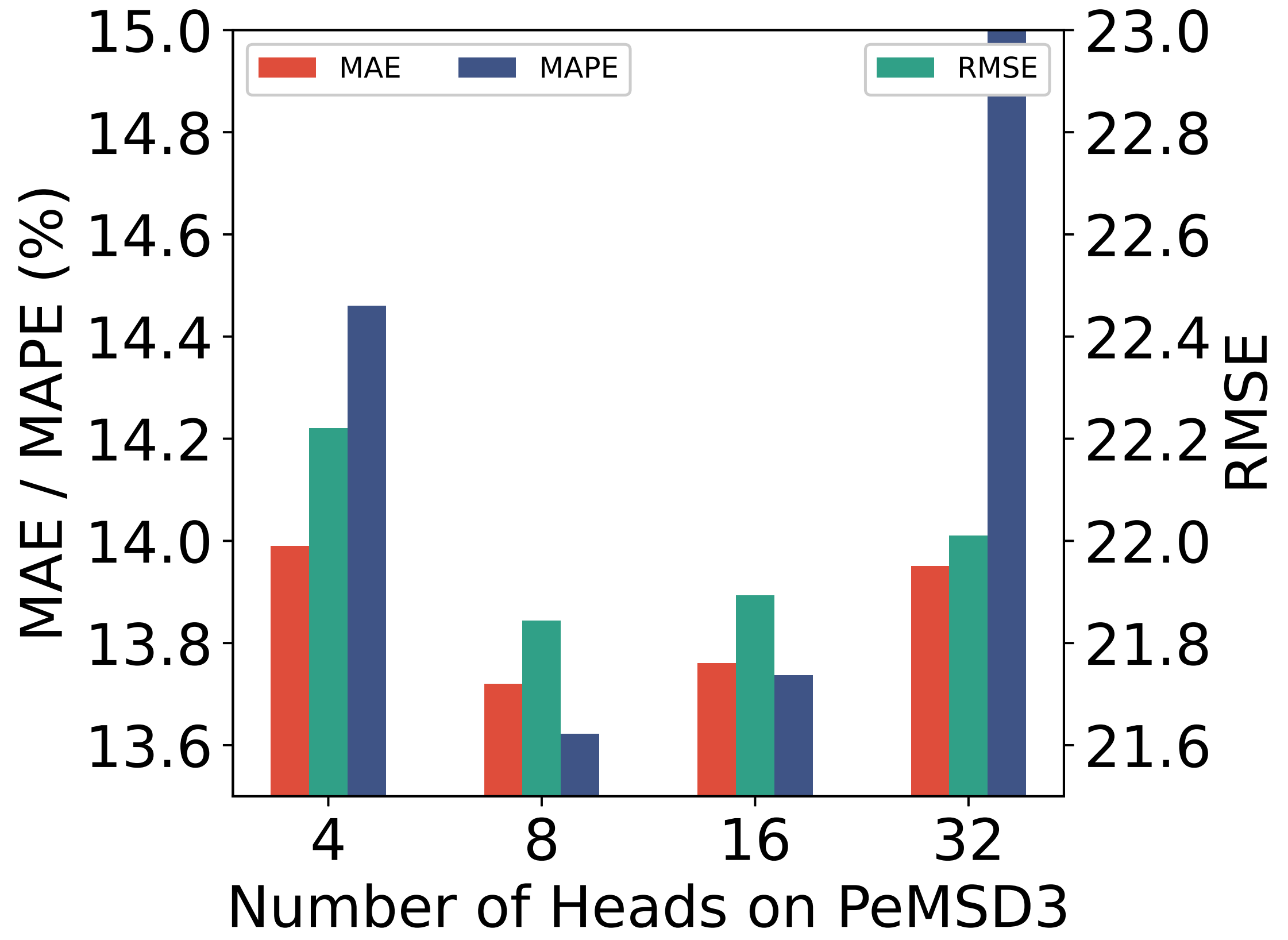}
        \end{minipage}
    }
    
    \caption{Influence of three parameters on four datasets.}
    \label{fig:hypar}
\end{figure*}

To study the influence of three parameters, i.e., the dimension of embedding $d_{model}$, the number of stacked encoder layers $L$, and the number of heads in the multi-head mechanism $h$, we conduct experiments with the range of these parameters mentioned above and $lr=1\times10^{-4}$ on all four datasets. Validation results are shown in Fig.~\ref{fig:hypar}, i.e., we can observe that $d_{model}=256$, $L=4$, $h=8$ best reflects the performance on PeMSD3.

    

\subsubsection{Effect of Input Embedding.} To investigate the effect of time embedding and positional embedding in our model, we evaluate five variants by (1) removing time embedding or positional embedding and (2) using fixed or learnable positional embedding. The results are shown in Table~\ref{tab:Ablation Experiment}. We observe that time embedding and learnable positional embedding could improve model performance, which demonstrating that our embedding mechanism effectively 
guide the Transformer encoder to learn the spatio-temporal correlation between vectors that represent historical traffic flow observations for each sensor.

\begin{table}
\centering
\caption{Effect of input embedding on PeMSD4}
\label{tab:Ablation Experiment}
\begin{tabular}{ccccc}
\hline
Time   Embedding & Postional Embedding & MAE & RMSE & MAPE (\%) \\ \hline
 &  & 22.59 & 35.48 & 16.92 \\
\checkmark &  & 21.65 & 34.80 & 15.21 \\
 & fixed & 22.14 & 34.54 & 18.48 \\
 & learnable & 19.38 & 30.97 & 15.95 \\
\checkmark & fixed & 18.55 & 30.32 & 14.16 \\ \hline
\checkmark & learnable & \textbf{18.49} & \textbf{30.29} & \textbf{13.10} \\ \hline
\end{tabular}
\end{table}

\section{Conclusion}
In this work, we propose a new fast traffic flow forecasting model: FPTN. In order to avoid processing vectors with dimensions larger than 800, the proposed FPTN decomposes the traffic flow data along the sensor dimension. Then, three kinds of embeddings are designed to sufficiently represent complicated spatio-temporal information in the decomposed sequences. Furthermore, Transformer encoder is utilized in FPTN to learn intricate spatio-temporal correlations concurrently and speed up forecast time. Extensive experiments on four real-world datasets show that FPTN achieves favorable performance compared with other state-of-the-art methods. x` Additionally, FPTN can also be used for other spatial-temporal forecasting tasks.

\bibliographystyle{splncs04}
\normalem

\bibliography{myref}
\end{document}